\documentclass{IEEEtran}
\usepackage{amsmath,epsfig,amssymb,bm,setspace,multirow,multicol,pifont,microtype}
\usepackage[utf8]{inputenc}
\usepackage[super]{nth}
\usepackage[ruled]{algorithm2e}
\usepackage{algorithmic}
\newcommand{\trans}[1]{{#1}^{\ensuremath{\mathsf{T}}}}
\newcommand{\scriptit}[1]{\scriptstyle{\mathsf{#1}}}

\newcommand{\xmark}{\ding{55}}

\title{Asynchronous Decentralized Distributed Training of Acoustic Models}
\author{Xiaodong Cui$^{1}$, Wei Zhang$^{1}$, Abdullah Kayi$^{1}$, Mingrui Liu$^{2}$, Ulrich Finkler$^{1}$, Brian Kingsbury$^{1}$, George Saon$^{1}$, David Kung$^{1}$ \\
$^{1}$IBM Research AI, IBM T. J. Watson Research Center, New York, USA \\
$^{2}$George Mason University, Virginia, USA}

\begin{document}

\maketitle

\begin{abstract}
Large-scale distributed training of deep acoustic models plays an important role in today's high-performance automatic speech recognition (ASR). In this paper we  investigate a variety of asynchronous decentralized distributed training strategies based on data parallel stochastic gradient descent (SGD) to show their superior performance over the commonly-used synchronous distributed training via allreduce, especially when dealing with large batch sizes. Specifically, we study three variants of asynchronous decentralized parallel SGD (ADPSGD), namely, fixed and randomized communication patterns on a ring as well as a delay-by-one scheme. We introduce a mathematical model of ADPSGD, give its theoretical convergence rate, and compare the empirical convergence behavior and straggler resilience properties of the three variants. Experiments are carried out on an IBM supercomputer for training deep long short-term memory (LSTM) acoustic models on the 2000-hour Switchboard dataset. Recognition and speedup performance of the proposed strategies are evaluated under various training configurations. We show that ADPSGD with fixed and randomized communication patterns cope well with slow learners. When learners are equally fast, ADPSGD with the delay-by-one strategy has the fastest convergence with large batches. In particular, using the delay-by-one strategy, we can train the acoustic model in less than 2 hours using 128 V100 GPUs with competitive word error rates.
\end{abstract}

\section{Introduction}
\label{sec:intro}

Automatic speech recognition (ASR) driven by deep learning \cite{LeCun_DLNature} has achieved unprecedented performance \cite{Hinton_DNNSPM}\cite{Xiong_ASRParity}\cite{Saon_2017HumanParity} due to improved acoustic models with deep architectures and large amounts of training data \cite{Parthasarathi_MillionhourASR}\cite{Parthasarathi_PetaAM}. It demands efficient training techniques to train the acoustic models in an acceptable period of time, which is crucial for ASR system optimization and product deployment in real-world applications. Distributed training under this condition is a desirable approach which has been shown in the machine learning community to significantly shorten the training time, notably \cite{Goyal_ImageNet1hour}\cite{Jia_ImageNet4mins}\cite{You_ImageNetinMins} in computer vision and \cite{You_BERTinMins} in natural language processing.\footnote{Accepted by IEEE/ACM Transactions on Audio, Speech and Language Processing. © 2021 IEEE.  Personal use of this material is permitted.  Permission from IEEE must be obtained for all other uses, in any current or future media, including reprinting/republishing this material for advertising or promotional purposes, creating new collective works, for resale or redistribution to servers or lists, or reuse of any copyrighted component of this work in other works.}

In ASR, distributed acoustic model training has been widely used \cite{Sainath_DNNBlueGene}\cite{Sak_DistSeq}\cite{Seide_ParaSGD}\cite{Strom_DSGD}. There is a broad variety of distributed training strategies \cite{Cui_DistASR} which, depending on their design on parallelism, synchronization mode and communication topology, fall into the following categories: model parallelism vs. data parallelism; synchronous vs. asynchronous; centralized vs. decentralized. Every design has had applications reported in the literature.

In \cite{Maas_ModelParal}, deep neural network (DNN) acoustic models were trained with model parallelism where parameters were distributed across multiple GPUs to handle acoustic models of large size. In \cite{Hannun_Deepspeech}, recurrent neural networks (RNNs) were split into two sub-models along the time dimension and distributed to two GPUs. Theoretical efficiency between model and data parallelism was compared in \cite{Seide_ParaSGD}. Compared to model parallelism, data parallelism distributes data in each mini-batch to multiple learners which comes naturally with stochastic gradient descent (SGD). It provides more flexibility in both algorithm and implementation and therefore has been the dominant approach in distributed training \cite{Povey_PDNN}\cite{Heigold_DistSeq}\cite{Chen_BMUF}\cite{Cong_KStepASR}\cite{Zhang_ADPSGDSWB}.

In terms of the synchronization mode, most existing strategies employ synchronous parallel stochastic gradient descent (PSGD) \cite{Seide_ParaSGD}\cite{Povey_PDNN}\cite{Chen_BMUF}\cite{Cong_KStepASR} where learners need to synchronize gradients or parameters for each mini-batch to update models. For instance, gradients are reduced in \cite{Seide_ParaSGD}\cite{Hannun_Deepspeech}\cite{Amodei_Deepspeech2}. In \cite{Seide_1bit}, which was one of the earliest research efforts on gradient compression in distributed machine learning \cite{Strom_DSGD}, 1-bit quantization of the gradient with error-compensation was introduced to substantially reduce the communication bandwidth. Other than gradient reduction, variants of K-step model averaging \cite{Zhou_KStepAvg} are also used. In \cite{Povey_PDNN}, periodic model averaging was implemented with natural gradient for SGD in distributed training. In \cite{Cong_KStepASR}, a minimal number of global model reductions was applied for each epoch to save the communication cost, allowing acoustic models to be trained on the 2000-hour Switchboard (SWB2000) dataset in around 3 hours on 96 GPUs. Blockwise model-update filtering (BMUF) proposed in \cite{Chen_BMUF} is another variant of K-step averaging where each learner carries out intra-block SGD in parallel. Instead of direct model averaging across learners, the global model is updated using block-level filtering with momentum. In the most recent work in \cite{Chen_BMUFAdam}, BMUF was extended from plain SGD to Adam \cite{Kingma_adam}. Gossip BMUF was introduced in \cite{Huang_GossipBMUF} in a decentralized framework where the entire model is divided into multiple components and each node randomly selects a few neighbors with which to communicate. In addition, discriminative synchronous distributed sequence training using natural gradient was reported in \cite{Haider_SeqNG} and later on extended to Hessian-free sequence training in \cite{Haider_SeqHFNG} to leverage second-order optimizers. Synchronous PSGD gives rise to good convergence but may suffer from the well-known ``straggler'' problem in which the slowest learner becomes the bottleneck of the training. This is a fairly common situation when one conducts distributed training on a cloud or a cluster competing with other users. If one or more nodes significantly lag behind their peers due to congested data traffic or resource overload, they will incur significant delay in the training. Oftentimes, one will find out that the model update has to wait a long time for the slowest learner to finish. Under this condition, synchronous PSGD will underperform. Compared to synchronous PSGD, asynchronous PSGD does not have the ``straggler'' issue and can automatically balance computational load \cite{Zhang_ADPSGDSWB}\cite{Cui_DistASR}. Representative work includes \cite{Heigold_DistDNN}\cite{Heigold_DistSeq}\cite{Zhang_ASGD} where each learner communicates with the parameter server in an asynchronous mode without waiting for other learners. For instance, Downpour SGD, which is a variant of asynchronous SGD, was used in \cite{Heigold_DistDNN} under the DistBelief framework \cite{Dean_LargeDistNN} for multilingual acoustic modeling. In this framework, learners push the updates to the parameter server without synchronization. The server is sharded across multiple machines and each shard is only responsible for updating part of the model. This strategy introduces asynchrony to both learners and shards of the parameter server. It has been shown to be robust to machine failure when dealing with large scale distributed training. However, asynchronous PSGD is harder to implement and debug, and its convergence may be significantly affected by the staleness problem, which results in poor recognition performance.

Communication in distributed training can take place in a centralized fashion with a parameter server or in a decentralized fashion. When there is a large number of learners, the communication cost to the parameter server can become large enough that it negates the desired gains from parallel training. In a decentralized algorithm, there is no central parameter server and all learners communicate among themselves \cite{Chen_BMUF}\cite{Zhang_ADPSGDSWB}\cite{Zhang_DDLASR}\cite{Amodei_Deepspeech2}.  Typically, learners form a ring to talk to each other via message passing protocols, avoiding the single-point communication bottleneck in a non-sharded parameter server, and is thus helpful for scaling out. Recently, decentralized strategies are becoming popular thanks to the availability of distributed software toolkits such as OpenMPI and the Nvidia Collective Communications Library (NCCL) \cite{NCCL}.

In this paper, we investigate an asynchronous decentralized PSGD (ADPSGD) framework where asynchrony is introduced to the decentralized training. We will show that ADPSGD can automatically balance the computational load with a moderate communication cost, which is friendly for heterogeneous computing environments and is suitable for scaling out distributed training. The theory of convergence of ADPSGD was first given in \cite{Lian_ADPSGD} which proved that under mild conditions the convergence rate of ADPSGD was on the same order as conventional mini-batch SGD for nonconvex objective functions. Since ADPSGD does not require synchronization among learners it is advantageous when dealing with slow learners. In addition, in ADPSGD the models used by each learner to evaluate the local gradients are heterogeneous, unlike the homogeneous models used in synchronous PSGD. These ``noisy'' models will be shown to help the model escape poor local optima, especially in large batch training. We will systematically investigate three strategies under this framework, namely, ADPSGD with a fixed communication pattern, ADPSGD with a randomized communication pattern, and ADPSGD with delay-by-one communication. We will analyze their convergence behaviors to consensus and compare their advantages and disadvantages. Their recognition and speedup performance will be evaluated on the 2000-hour Switchboard dataset under various computing configurations followed by a general discussion on how to leverage synchronous and asynchronous PSGD given different training environments. Notably, we will show that when there are no stragglers, the delay-by-one variant yields the best scaling performance with competitive recognition performance.\footnote{This manuscript is a significant extension of the work presented at ICASSP 2020 \cite{Zhang_ADPSGDRAND}.}

The remainder of the paper is organized as follows. Section \ref{sec:adpsgd} is devoted to the theory of ADPSGD, providing a mathematical model with a non-asymptotic ergodic convergence rate. The implementation of ADPSGD is presented in Section \ref{sec:real}, in which three ADPSGD strategies are introduced and analyzed. Experimental results on SWB2000 are presented in Section \ref{sec:exp}, followed by a discussion in Section \ref{sec:disc}. We conclude the paper with a summary in Section \ref{sec:sum}.

\section{ADPSGD}
\label{sec:adpsgd}

ADPSGD was proposed in \cite{Lian_ADPSGD} and was first used for large-scale acoustic model training in \cite{Zhang_ADPSGDSWB}. In this section, we introduce a mathematical model, review its convergence rate, and investigate three ADPSGD variants.

\subsection{Mathematical Formulation}
\label{sec:model}

Consider the following distributed optimization problem over $L$ learners
\begin{align}
    \min _{w} \  F(w) = \frac{1}{L}\sum_{i=1}^{L}\mathbb{E}_{\xi \thicksim \mathcal{D}_{i}}[f(w,\xi)]
\end{align}
where $w$ is the parameters to be optimized. For each learner $i$, $F_{i}(w)\triangleq \mathbb{E}_{\xi \thicksim \mathcal{D}_{i}}[f(w,\xi)]$ is the local loss function and $\mathcal{D}_{i}$ is the local data distribution. $\xi \thicksim \mathcal{D}_{i}$ is a random variable sampling data from distribution $\mathcal{D}_{i}$. We further assume all learners can access the whole training data set, which means that samples on each learner are independent and identically distributed (i.i.d.).  All $\mathcal{D}_{i}$ are identical to the global empirical data distribution. When the model is a neural network, $w$ is the weights of the network.

The optimization is carried out using mini-batch PSGD, the pseudo-code of which is given in Algorithm \ref{alg:adpsgd}.
\begin{algorithm}
    \caption{ADPSGD}
    \label{alg:adpsgd}
     \textbf{Input:} Same initial local model $w^{(l)}_{0}=w_{0}$; number of learners $L$; local batch size $M$; total number of iterations $K$; learning rate schedule $\{\alpha_{k}\}$.

     \For {$k = 1 : K$}
     {
         \tcp{for each learner $l$}
         \textbf{Run concurrently}:

         \ \ \ \ \textbf{Gradient computation}

         \ \ \ \ \ \ \ \ Sample a mini-batch of size $M$;

         \ \ \ \ \ \ \ \ Use the current local model $w^{(l)}_{k}$ to compute

         \ \ \ \ \ \ \ \ gradient $g_{k}=\frac{1}{M}\sum_{m=1}^{M} \nabla f(w^{(l)}_{k};\xi^{(1)}_{k,m})$;

         \ \ \ \ \textbf{Model averaging}

         \ \ \ \ \ \ \ \ Select a doubly stochastic matrix $\mathbf{T}_{k}=[t^{k}_{ij}]$;

         \ \ \ \ \ \ \ \ Average local model with models from other

         \ \ \ \ \ \ \ \ learners $w^{(l)}_{k+\frac{1}{2}} = \sum_{j=1}^{L}t^{k}_{lj}w^{(j)}_{k}$; \medskip

         \textbf{Local model update}

         \ \ \ \ \ \ \ \ $w^{(l)}_{k+1} = w^{(l)}_{k+\frac{1}{2}} - \alpha_{k} g^{(l)}_{k}$; \medskip

     }

     Output the final model as the average of models from all learners  $w_{K} = \frac{1}{L}\sum_{l=1}^{L}w^{(l)}_{K}$. \medskip
\end{algorithm}

The $L$ learners form a communication graph. For each learner, when the mini-batch gradient is being computed, model averaging with other selected learners on the graph is also conducted simultaneously. Once the gradient computation is over, the local model on the learner will be updated with the averaged model without synchronizing with other learners. Since the model used to evaluate the gradient may be different from the one to be updated, as it may have been changed after averaging with other learners while the gradient is being computed, staleness occurs.

In general, the model update of ADPSGD with $L$ learners can be written as
\begin{align}
    \mathbf{W}_{k+1} = \mathbf{W}_{k}\mathbf{T}_{k} - \alpha_{k}g(\mathbf{\Phi}_{k},\bm{\xi}_{k})  \label{eqn:dsgd}
\end{align}
where, for $l = 1, \dots, L$ and local batch size $M$, $\mathbf{W}_{k} = [w^{(1)}_{k}, \dots, w^{(l)}_{k}, \dots, w^{(L)}_{k}]$ is a matrix with each column containing model parameters in each learner $l$ at iteration $k$; $\mathbf{T}_{k}$ is a doubly stochastic mixing matrix\footnote{A doubly stochastic matrix is a square matrix of non-negative real numbers. The sums of each row and column of the matrix are 1.} specifying model averaging between learners on the graph at iteration $k$; $\mathbf{\Phi}_{k} = [\hat{w}^{(1)}_{k}, \dots, \hat{w}^{(l)}_{k}, \dots, \hat{w}^{(L)}_{k}]$ is a matrix with each column containing model parameters used for computing the gradient in each learner $l$ at iteration $k$; $\bm{\xi}_{k} = [\xi^{(1)}_{k}, \dots, \xi^{(l)}_{k}, \dots, \xi^{(L)}_{k}]$ is a matrix with each column containing indexing random variables for mini-batch samples used for computing the gradient in each learner $l$ at iteration $k$;  $g(\mathbf{\Phi}_{k},\bm{\xi}_{k}) = [\frac{1}{M}\sum_{m=1}^{M} \nabla f(\hat{w}^{(1)}_{k};\xi^{(1)}_{k,m}), \dots, \frac{1}{M}\sum_{m=1}^{M} \nabla f(\hat{w}^{(L)}_{k};\xi^{(L)}_{k,m})]$ is a matrix with each column containing the gradient computed in each learner $l$ at iteration $k$. $\hat{w}^{(l)}_{k} = w^{(l)}_{k-\tau^{(l)}_{k}}$ where $\tau^{(l)}_{k}$ is the staleness. In this paper, asynchrony is referred to as model staleness or, more broadly, heterogeneous models on learners to evaluate local gradients.

It was proved in \cite{Lian_ADPSGD} (Corollary 2) that for nonconvex objective function $f(w)$, under mild conditions, with an appropriately chosen learning rate and bounded staleness, when data on learners is i.i.d and $K$ is sufficiently large, we have:
\begin{align}
 & \frac{\sum_{k=0}^{K-1}\mathbb{E}\parallel \nabla f(\frac{1}{L}\sum_{l}w^{(l)}_{k})\parallel^{2}}{K}  \nonumber \\
 \leq & \frac{20(f(w_{0})-f^{*})\mu}{K} + \frac{2(f(w_{0})-f^{*}+\mu)\sigma}{\sqrt{MK}} \label{eqn:rate}
\end{align}
where $\mu$ is the Lipschitz constant of the gradient, $\sigma$ the upper bound of the variance of the unbiased gradient and $f^{*}$ the global optimum of the objective function $f(w)$.

Eq.~\ref{eqn:rate} indicates that ADPSGD converges with an ergodic convergence rate of $\mathcal{O}(\frac{1}{K})+\mathcal{O}(\frac{1}{\sqrt{MK}})$. When $K$ is large, the second term dominates, which gives a rate of $\mathcal{O}(\frac{1}{\sqrt{MK}})$. This is on the same order as conventional SGD on nonconvex functions \cite{Ghadimi_SGDrate}, which is $\mathcal{O}(\frac{1}{\sqrt{K}})$, but with a linear speedup in batch size $M$.

\section{Realization}
\label{sec:real}

\begin{figure}[htb]
  \centering
  \centerline{\includegraphics[width=7cm, height=7.5cm]{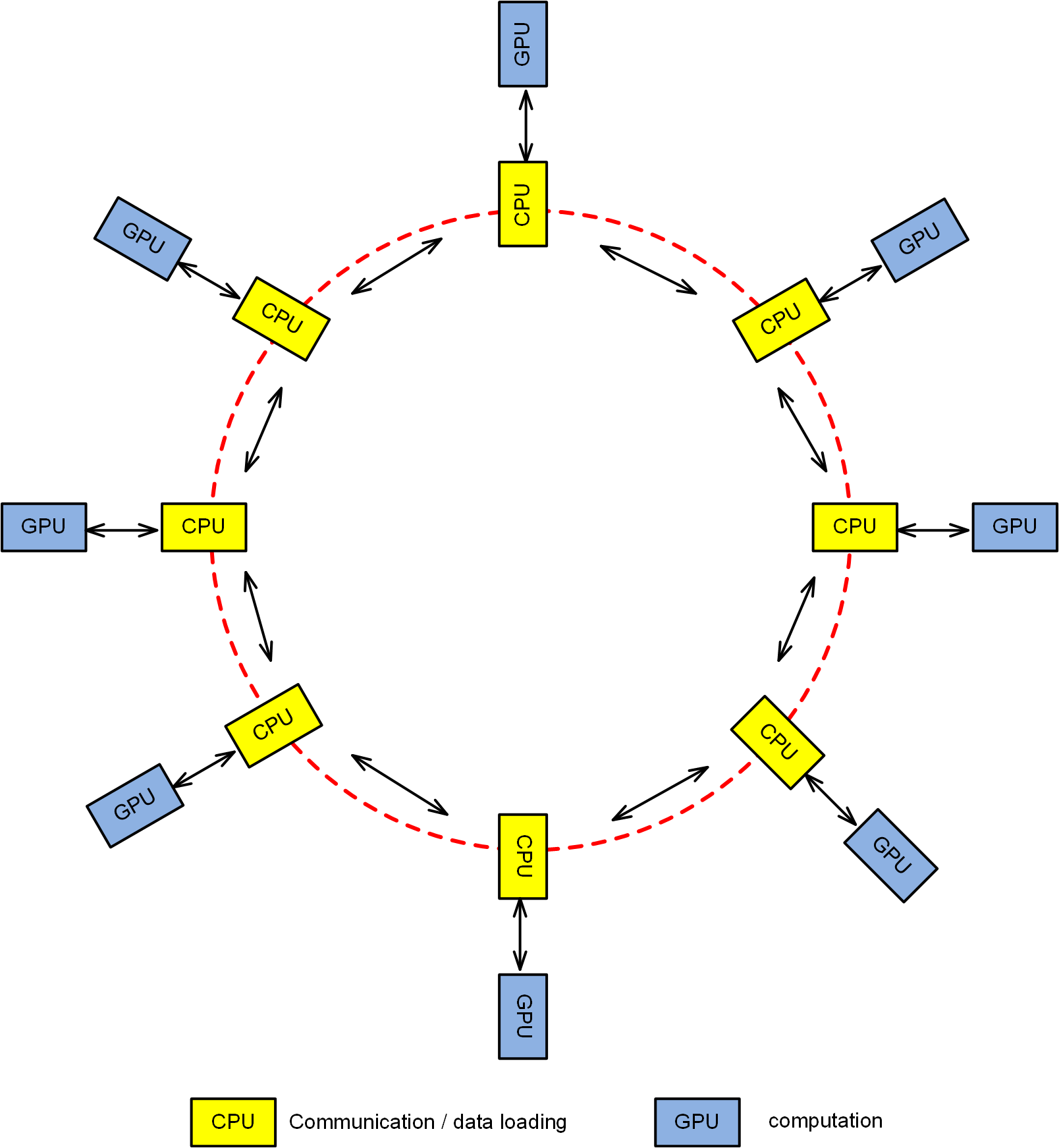}}
  \caption{An illustration of implementation of ADPSGD. All learners form a ring. Each learner has local gradient computation carried out on GPU and model averaging with other learners and data loading on CPU. The two processes are carried out simultaneously.}\label{fig:ring}
\end{figure}

ADPSGD is usually realized on a ring-based communication graph as illustrated in Fig.~\ref{fig:ring}. Each learner has a GPU and CPU. The gradient evaluation, the second term of the right-hand-side (RHS) of Eq.~\ref{eqn:dsgd}, which is computationally heavy, is put on the GPU while model averaging and data loading, which are communication heavy, are put on the CPU. The two processes are carried out concurrently to achieve maximum overlap between the two operations. Every learner pushes the gradient from GPU to CPU as soon as the gradient computation is finished, without waiting for other learners, and starts another round of gradient computation and model averaging.

\subsection{Allreduce}
\label{sec:allreduce}

In the following, we investigate three realizations of ADPSGD under various mixing strategies and compare their performance in terms of convergence speed and communication cost. Before diving into the details of these strategies it is helpful to note that Eq.~\ref{eqn:dsgd} can describe the commonly-used synchronous decentralized PSGD (SDPSGD) via allreduce as a special case by introducing synchronization of appropriate order and a chosen mixing matrix $\mathbf{T}_{k}$.  Allreduce applies a sum reduction operation on gradients or models from all learners and sends the result back to each learner. Each learner thus receives the sum of gradients or models and then divides it by the number of learners locally to get identical gradient or model averages. Suppose instead of running simultaneously, the model averaging (the first term of RHS of Eq.~\ref{eqn:dsgd}) takes place after all learners finish their local gradient computation and local model update. Choosing the mixing matrix to be
\begin{align}
      \mathbf{T}_{k} = \frac{\mathbf{1}_{L}\trans{\mathbf{1}}_{L}}{L}, \ \ \ \  \mathbf{1}_{L} = \trans{[1,\cdots,1]} \in \mathbb{R}^{L\times 1},
\end{align}
will result in identical local models, which is the global average $w_{k+1} = \frac{1}{L}\sum_{i=1}^{L}w^{(l)}_{k}$. For plain SGD in Eq.~\ref{eqn:dsgd}, one-step model averaging and gradient averaging are equivalent \cite{Cui_DistASR}, so it also amounts to allreduce on gradient under this condition. Algorithm \ref{alg:sdpsgd_iter} gives the pseudocode of this implementation.

\begin{algorithm}
\caption{SDPSGD on a learner in iteration $k$}
\label{alg:sdpsgd_iter}

$w_{k} \leftarrow \text{pull\_model\_from\_ring}()$; \tcp{sync.}

$g_{k} \leftarrow \text{eval\_local\_gradient}(w_{k})$;

$w_{k+\frac{1}{2}} \leftarrow \text{update\_local\_model}(w_{k},g_{k})$;

$\text{push\_model\_to\_ring}(w_{k+\frac{1}{2}})$;

$w_{k+1} \leftarrow \text{allreduce\_on\_ring}(w_{k+\frac{1}{2}})$;
\end{algorithm}

In generic decentralized distributed training, one typically has to deal with two stochastic processes simultaneously. One is all learners trying to converge to consensus, and the other is the averaged model trying to converge to a local optimum. If the convergence of the two stochastic processes is approximately on the same order, then faster convergence of learners to consensus will lead to faster overall convergence. In allreduce the consensus is reached by one-step averaging across all the learners on the ring. The consensus of learners relies on the structure of the doubly stochastic mixing matrix. In fact, for all doubly stochastic mixing matrices $\mathbf{T}$,
\begin{align}
    \mathbf{T}^{t} \rightarrow \frac{\mathbf{1}_{L}\trans{\mathbf{1}}_{L}}{L},  \ \ \ \  t \rightarrow \infty .
\end{align}
Since the largest eigenvalue of a doubly stochastic matrix is always 1, the speed of convergence is controlled by the second largest eigenvalue of $\mathbf{T}$ \cite{Lian_DecentSGD}:
\begin{align}
\left\|\mathbf{T}^{t}-\frac{\mathbf{1}_{L}\trans{\mathbf{1}}_{L}}{L}\right\|_2\leq \hat{\lambda}^{t} \label{eqn:rho}.
\end{align}
where
\begin{align}
    \hat{\lambda} = \max_{i=2,\dots,L}|\lambda_{i}(\mathbf{T})|.
\end{align}
The difference between the two largest eigenvalues
\begin{align}
   \rho = 1 - \hat{\lambda}
\end{align}
is referred to as the spectral gap.

\subsection{ADPSGD with Fixed Mixing (ADPSGD-FM)}
\label{sec:adpsgd_fix}

Algorithm \ref{alg:adpsgd_iter} gives the pseudocode for one ADPSGD iteration on one learner.

\begin{algorithm}
\caption{ADPSGD on a learner in iteration $k$}
\label{alg:adpsgd_iter}

\textbf{Run concurrently}:

\ \ \ \ $g_{k} \leftarrow \text{eval\_local\_gradient}(w_{k})$;

\ \ \ \ $w_{k+\frac{1}{2}} \leftarrow \text{model\_averaging\_on\_ring}(w_{k})$;

$w_{k+\frac{1}{2}} \leftarrow \text{pull\_model\_from\_ring}()$;

$w_{k+1} \leftarrow \text{update\_local\_model}(w_{k+\frac{1}{2}},g_{k})$;

$\text{push\_model\_to\_ring}(w_{k+1})$;
\end{algorithm}

Consider the following fixed mixing (FM) matrix in model averaging:
\begin{align}
\mathbf{T}^{f}_{k} = \begin{bmatrix}
    \frac{1}{3} & \frac{1}{3} &    0        &    0        &     0     &     0         &   \frac{1}{3}  \\
    \frac{1}{3} & \frac{1}{3} & \frac{1}{3} &    0        &     0     &     0         &       0        \\
    0           & \frac{1}{3} & \frac{1}{3} & \frac{1}{3} &     0     &     0         &       0        \\
    0           &     0       & \frac{1}{3} & \frac{1}{3} & \frac{1}{3} &     0     &     0            \\
      \cdots    &   \cdots    &  \cdots     &  \cdots     &   \cdots  & \cdots        &   \cdots  \\
    \frac{1}{3} &    0        &    0        &    0        &     0     &  \frac{1}{3}  &  \frac{1}{3}
\end{bmatrix}. \label{eqn:matrix_fm}
\end{align}

With this mixing matrix, every learner only communicates with its immediate left and right neighbors on the ring. Therefore, compared to allreduce in which communication has to be conducted among all the learners, this mixing strategy can significantly reduce the communication cost, especially when there is a large number of learners. In Appendix~\ref{app:specgap_fm} it is shown that, given the circulant structure of $\mathbf{T}^{f}_{k}$, the second largest eigenvalue is
\begin{align}
    \hat{\lambda} = \frac{1}{3} + \frac{2}{3}\cos\left(\frac{2\pi}{L}\right).  \label{eqn:trirho}
\end{align}
Therefore, we have
\begin{align}
\left\|\mathbf{T}^{f}_{1}\mathbf{T}^{f}_{2}\cdots\mathbf{T}^{f}_{k}-\frac{\mathbf{1}_{L}\trans{\mathbf{1}}_{L}}{L}\right\|_2\leq \left(\frac{1}{3} + \frac{2}{3}\cos\left(\frac{2\pi}{L}\right)\right)^{k}. \label{eqn:rate_fm}
\end{align}
Eq.~\ref{eqn:rate_fm} indicates that the convergence of ADPSGD-FM to consensus among learners is affected by the number of learners $L$. As $L$ becomes large, the spectral gap $\rho$ will be close to 0, which will slow down the convergence. Intuitively, when there are many learners, if each learner only talks to its left and right neighbors, information takes a longer time to diffuse on the ring to reach consensus.

\subsection{ADPSGD with Random Mixing (ADPSGD-RM)}
\label{sec:adpsgd_rand}

Instead of only communicating with left and right neighbors, random mixing (RM) communicates with two randomly selected learners for model averaging in each iteration. Its implementation still follows Algorithm \ref{alg:adpsgd_iter}, except that FM is replaced by RM in the model averaging step.

Let's shuffle indices of the $L$ learners on the ring:
\begin{align}
    [1,2,\ldots,L] \rightarrow [\sigma(1), \sigma(2), \ldots, \sigma{(L)}]
\end{align}
where $\sigma(\cdot)$ is a random permutation of the set $\{1,\dots, L\}$. Each learner averages its models with its left and right neighbors in the mapped indices, which gives the mixing matrix
\begin{align}
    \mathbf{T}^{r}_{k} = \trans{\mathbf{P}}_{k}\mathbf{T}^{f}_{k}\mathbf{P}_{k}   \label{eqn:randperm}
\end{align}
where $\mathbf{P}_{k}$ is a random permutation matrix. It is shown in Appendix \ref{app:specgap_rm} that
\begin{align}
\mathbf{E}\left\|\mathbf{T}^{r}_{1}\mathbf{T}^{r}_{2}\cdots\mathbf{T}^{r}_{k}-\frac{\mathbf{1}_{L}\trans{\mathbf{1}}_{L}}{L}\right\|_2 \leq \frac{\sqrt{L-1}}{(\sqrt{3})^{k}}. \label{eqn:rate_rm}
\end{align}
Comparing with Eq.~\ref{eqn:rate_fm}, we can see that random mixing converges to consensus much faster than fixed mixing. Fig.~\ref{fig:decay} illustrates the rate of decay in the RHS of Eq.\ref{eqn:rate_fm} and Eq.\ref{eqn:rate_rm} with $L\!=\!16$, $32$ and $64$. Meanwhile, each learner still only communicates with two other learners at a time. So the communication cost is similar to that of fixed mixing, bar some overhead required to connect non-adjacent learners. This additional communication cost consists of two parts. First of all, the random neighbors need to be selected on the ring. This is accomplished by random permutation implemented by the Fisher-Yates shuffle~\cite{Durstenfeld_RandPerm}, which has a complexity of $\mathcal{O}(L)$ and thus is negligible compared to data loading and model communication. Second, once the neighbors are chosen, the communication takes place among the learners. If the learners are all located on the same node, fixed and random mixing have the same cost in the case of single-socket CPUs as they both use the main memory bandwidth. But, in the case of multi-socket CPUs, fixed mixing is faster, but both fixed and random mixing are likely bounded by the inter-socket communication bandwidth and are thus unlikely to differ much. If the learners are located on different nodes, random mixing will be bounded by the inter-node communication bandwidth, which is slower than the intra-node bandwidth. However, in typical high performance computing (HPC) settings, nodes in the same rack are connected to the same network switch in a ``fat-tree'' fashion. The communication takes place first from a learner to a switch and then from a switch to a learner. In general, the additional communication cost incurred by RM over FM is quite tolerable.

\subsection{ADPSGD with Delay-by-One (ADPSGD-D1D)}
\label{sec:adpsgd_d1d}

In the delay-by-one (D1D) strategy, which we will find to significantly outperform RM and FM in the absence of stragglers, model averaging and gradient computation are carried out concurrently. The model averaging is realized with allreduce. Therefore,
\begin{align}
\mathbf{T}^{d}_{k} = \frac{\mathbf{1}_{L}\trans{\mathbf{1}}_{L}}{L}.
\end{align}
When the two operations are finished, they are synchronized to update the model before the next iteration. Therefore, the two terms on the RHS of Eq.~\ref{eqn:dsgd} are realized as $\mathbf{W}_{k}\mathbf{T}^{d}_{k}$ and $g(\mathbf{\Phi}_{k},\bm{\xi}_{k})$ where
\begin{align}
    \mathbf{\Phi}_{k} = [w^{(1)}_{k-1}, \dots, w^{(l)}_{k-1}, \dots, w^{(L)}_{k-1}].  \label{eqn:d1d_phi}
\end{align}
Because of the overlap between gradient computation and model allreduce, the local model used to evaluate gradients lags the allreduced model by precisely one iteration, thus the name delay-by-one. From Eq. \ref{eqn:rho}, D1D achieves an upper bound on convergence speed in the ADPSGD setting because the background model averaging is achieved by a global allreduce. The implementation details are given in Algorithm \ref{alg:d1d_iter}.

\begin{algorithm}
\caption{D1D on a learner in iteration $k$}
\label{alg:d1d_iter}

\textbf{Run concurrently}:

\ \ \ \ $g_{k} \leftarrow \text{eval\_local\_gradient}(w_{k})$;

\ \ \ \ $w_{k+\frac{1}{2}} \leftarrow \text{model\_average\_on\_ring}(w_{k})$;

$w_{k+\frac{1}{2}} \leftarrow \text{pull\_model\_from\_ring}()$; \tcp{sync.}

$w_{k+1} \leftarrow \text{update\_local\_model}(w_{k+\frac{1}{2}},g_{k})$;

$\text{push\_model\_to\_ring}(w_{k+1})$;
\end{algorithm}

Since D1D employs allreduce, it sacrifices asynchrony and faces the same ``straggler'' issue as other synchronous training approaches. However, from Algorithm \ref{alg:d1d_iter} we can see that learners in D1D still concurrently conduct local gradient computation and model averaging. But, to pull the averaged model from the ring, each learner has to wait for the model averaging to finish. More importantly, D1D use non-identical models to evaluate local gradients in Eq. \ref{eqn:d1d_phi}. This results in heterogeneous local models due to the delay-by-one asynchrony. This is the major difference from other allreduce-based synchronous training methods in which homogeneous models are used to evaluate the local gradient on each learner. This difference is important as it introduces noise which can prevent learning from being trapped early in a poor local minimum. This property will be shown later to be very helpful for scaling out with a large batch size and aggressive learning rate.

In \cite{Seide_1bit}, a double-buffering technique was proposed to improve the concurrency of gradient computation and communication. Double-buffering bears resemblance to D1D in the sense that it requires synchronization based on allreduce and improves concurrency by introducing deterministically controlled staleness which is 0.5 due to the half-batch buffering. However, double-buffering is essentially still a conventional synchronous PSGD technique where homogeneous models are used by all learners to compute their local gradients.

\section{Experimental Results}
\label{sec:exp}

The experiments are conducted on the SWB2000 dataset \cite{Godfrey_SWB}\cite{Cieri_Fisher} consisting of 1,975 hours of audio from which 10 hours are held out as the validation set. Test sets include a 2.1-hour switchboard (SWB) set and 1.6-hour call-home (CH) set. The acoustic model is a hybrid LSTM deep neural network-hidden Markov model (DNN-HMM). There are 6 bi-directional layers in the model with 1,024 cells in each layer (512 cells in each direction). There is a linear projection layer of 256 hidden units between the topmost LSTM layer and the softmax output layer.  There are 32,000 output units in the softmax output layer corresponding to context-dependent HMM states. The LSTM is unrolled over 21 frames and trained with non-overlapping feature subsequences of that length. The dimensionality of the input features is 260 which is a fusion of 40-dim FMLLR, 100-dim i-vector and 40-dim logmel with delta and double delta coefficients. The language model is built using publicly available training data from a broad variety of sources. There are 36M 4-grams built on a vocabulary of 85K words. Training is carried out using cross-entropy (CE) loss. For a mini-batch of size $M$ 21-frame segments, the input tensor is of size $M\!\times\!260\!\times\!21$.

Distributed training experiments are carried out on an IBM supercomputer having a similar architecture and hardware specs as Summit \cite{summit}, the fastest supercomputer in the United States and the second fastest supercomputer in the world. It is based on IBM POWER System AC922 nodes with IBM POWER9 CPUs and NVIDIA Volta V100 GPUs, all connected together with Nvidia’s high-speed NVLink dual links totaling 50GB/s bandwidth in each direction. Each node contains 22 cores, 512GB of DDR4 memory, 96GB of High Bandwidth Memory (HBM2) for use by the accelerators and is equipped with 6 GPUs. Nodes are connected with Mellanox EDR 100G Infiniband interconnect technology. Each each node has a combined network bandwidth of 25GB/s and is equipped with 500GB NVME storage. Allreduce-based SDPSGD and ADPSGD-D1D are implemented using NCCL. Since ADPSGD-FM and ADPSGD-RM involve more generic partial model averaging on a communication graph, they are implemented using MPI.

The baseline is established by training the acoustic model using SGD on a single V100 GPU without parallelization. The initial learning rate is 0.1 which is annealed by $\frac{1}{\sqrt{2}}$ every epoch after the \nth{10} epoch. The training finishes after 16 epochs. Table \ref{tab:baseline} presents the word error rates (WERs) and training time with batch sizes of 128, 256 and 512 segments on a single V100 GPU.

\begin{table}[htb]
\centering
\begin{tabular}{c|c|c|c}\hline
\multirow{2}{*}{batch size} & \multicolumn{2}{c|}{WER(\%)}  & \multirow{2}{*}{Time(h)}  \\ \cline{2-3}
                            &   SWB     &     CH            &                            \\ \hline
         128                &   7.5     &    13.0           &       121.96               \\ \hline
         256                &   7.5     &    13.0           &        99.00               \\ \hline
         512                &   7.5     &    13.1           &        82.06               \\ \hline
\end{tabular}
\vspace{0.3cm}
\caption{WERs and training time for baseline using single V100 GPU.}\label{tab:baseline}
\end{table}

\subsection{Convergence Speed}
\label{sec:convergece}

\begin{figure}[htb]
  \centering
  \centerline{\includegraphics[width=8cm, height=6cm]{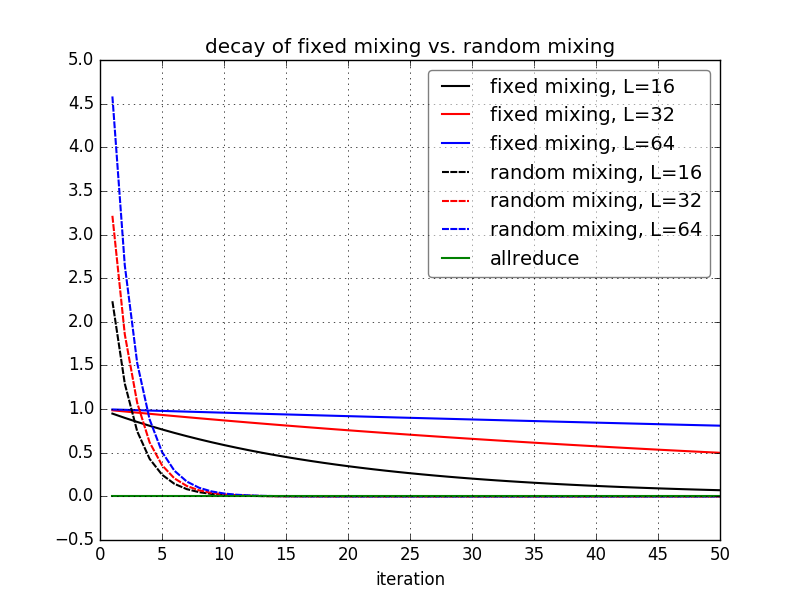}}
  \caption{Convergence speed to consensus under fixed mixing and random mixing with various numbers of learners. The Y-axis represents the bound over iterations on the $L_{2}$ distance between the product of mixing matrices and the consensus, as shown in Eq.\ref{eqn:rate_fm} and Eq.\ref{eqn:rate_rm}}\label{fig:decay}
\end{figure}

\begin{figure}[htb]
  \centering
  \centerline{\includegraphics[width=9.5cm, height=3.1cm]{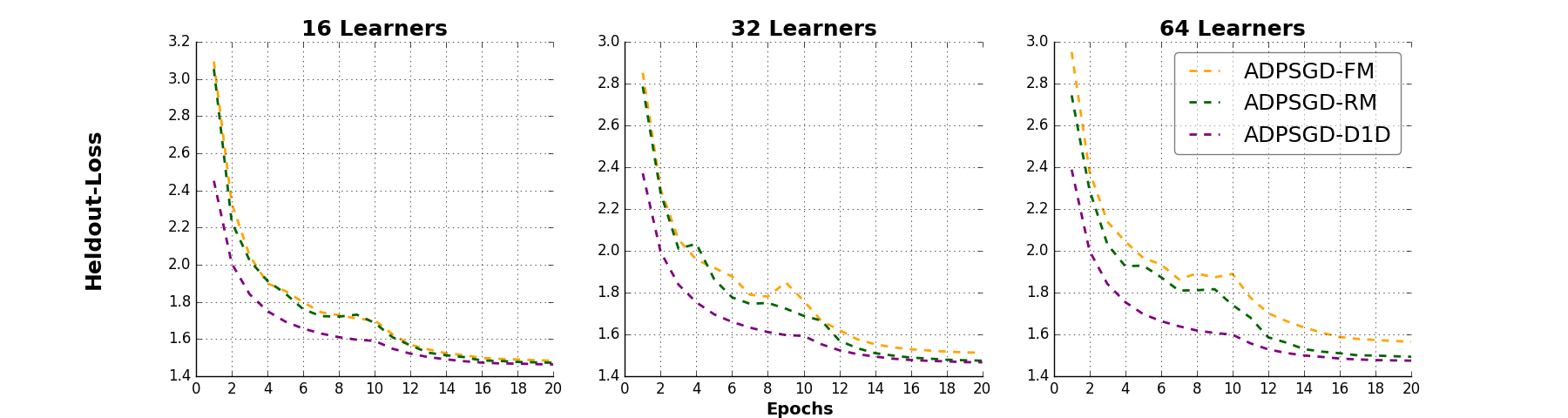}}
  \caption{Heldout loss over epochs under the three ADPSGD strategies -- fixed mixing (ADPSGD-FM), random mixing (ADPSGD-RM) and delay-by-one (ADPSGD-D1D) using 16, 32 and 64 learners. }\label{fig:converge_comp}
\end{figure}

The convergence speed of the ADPSGD strategies has significant dependence on the degree of consensus among the learners. Fig.~\ref{fig:decay} illustrates the bound on the $L_{2}$ distance between the product of mixing matrices and the consensus as a function of iterations, as shown in Eq.\ref{eqn:rate_fm} and Eq.\ref{eqn:rate_rm}. The curves show how the speed at which global consensus is reached depends on the different spectral gaps of the three strategies. D1D has the fastest speed as it is essentially an allreduce which achieves global consensus in one step. Random mixing approaches consensus much faster than fixed mixing, especially when the number of learners is large. The theory is also clearly verified in the experiments indicated in Fig.~\ref{fig:converge_comp}. The figures show the loss on the heldout set of the three strategies using 16, 32 and 64 learners over 20 epochs of training. The total batch size is 8,192 segments. For fixed mixing, the convergence slows down when the number of learners goes from 16 to 64. Given the same number of learners, D1D always achieves the fastest convergence, while random mixing converges faster than fixed mixing. The difference in convergence speed between the fixed and random mixing is not obvious when using 16 learners, but becomes evident when using 64 learners.

\subsection{Large batch size}
\label{sec:largebatch}

\begin{figure}[htb]
  \centering
  \centerline{\includegraphics[width=8cm, height=6cm]{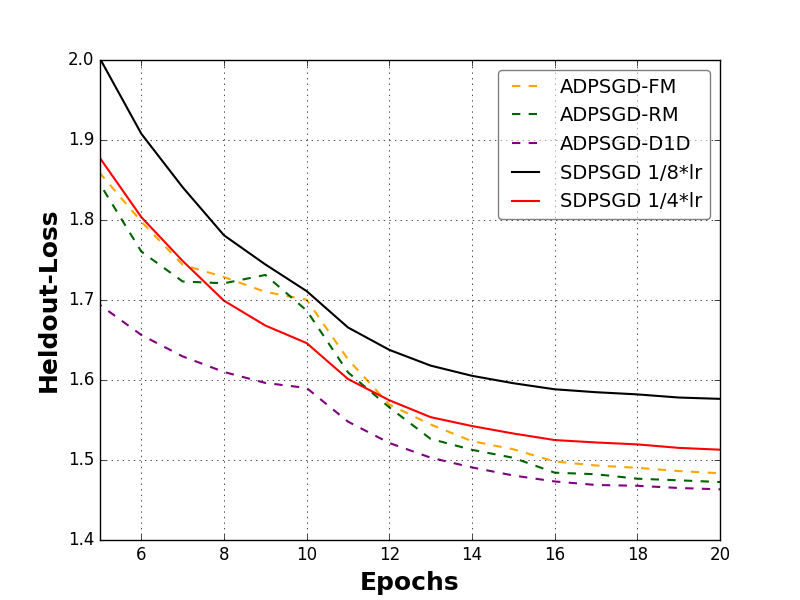}}
  \caption{Convergence of three ADPSGD variants and synchronous PSGD using 8,192-segment batches and 16 learners. Synchronous PSGD will not converge following the same learning rate schedule as the ADPSGD variants or scaling the learning rate by $\frac{1}{2}$. Scaling the learning rate by $\frac{1}{4}$ and $\frac{1}{8}$ will converge, but to poorer local optima with higher heldout loss.}\label{fig:lbs_convergence}
\end{figure}

A large batch size plays a crucial role in scaling out distributed training under data parallelism because there needs to be enough computation per round of communication to achieve faster training. For conventional allreduce-based synchronous PSGD, it is often observed that too large a batch size may result in either early divergence or a poor local optimum \cite{Goyal_ImageNet1hour}\cite{You_BERTinMins}\cite{Kumar_TPU}\cite{Zhang_study}. We show that the ADPSGD variants we investigate are more friendly to large batch sizes compared to synchronous PSGD and thus more advantageous in large scale distributed training. In \cite{Seide_ParaSGD}\cite{Goyal_ImageNet1hour}, large batch training uses a large learning rate. A rule of thumb is that the learning rate should be scaled approximately in proportion to the batch size. To avoid divergence with a very large learning rate, a warmup process is employed in the early stage of the training. Similar observations have been made in other work \cite{You_BERTinMins}, and this has become a widely used practice in large scale distributed training. In this work, we also follow a similar procedure. Fig.~\ref{fig:lbs_convergence} compares the convergence curves of ADPSGD under fixed mixing (ADPSGD-FM), random mixing (ADPSGD-RM) and delay-by-one (ADPSGD-D1D) to the convergence curves of SDPSGD using 8,192-segment batches and 16 learners. All three ADPSGD variants use the same learning rate schedule. For 8,192-segment batches we use a learning rate of 3.2, which is 32 times larger than the learning rate used for 256-segment batches. We start from 0.32, employ a linear warmup in the first 10 epochs to reach 3.2, and then anneal the learning rate by $\frac{1}{\sqrt{2}}$ in each epoch afterwards.

Synchronous PSGD following the same learning rate schedule simply diverges. To make it converge, we scale the learning rate by $\frac{1}{2}$, $\frac{1}{4}$ and $\frac{1}{8}$. It turns out that even scaling by $\frac{1}{2}$ still leads to divergence, while scaling by $\frac{1}{4}$ and $\frac{1}{8}$ converges, but to a much higher held-out loss than the ADPSGD variants. This is indicated by WERs in Table \ref{tab:lbs_wer}.

\begin{table}[hbt]
 \centering
\begin{tabular}{c|c|c|c|c|c|c}\hline
\multirow{2}{*}{} & \multirow{2}{*}{Baseline}  & \multicolumn{3}{c|}{ADPSGD} & \multicolumn{2}{c}{SDPSGD}  \\ \cline{3-7}
                  &           &    FM   &   RM   &    D1D   &  1/4*lr   &  1/8*lr   \\ \hline\hline
SWB               & 7.5       & 7.6         & 7.6         & 7.4          &   7.8                  &  8.3                   \\ \hline
CH                & 13.0      & 13.2        & 13.1        & 13.3         &   13.5                 &  14.3   \\ \hline
\end{tabular}
\vspace{0.3cm}
\caption{WERs of three ADPSGD strategies and synchronous PSGD using 8,192-segment batches and 16 learners.}\label{tab:lbs_wer}
\end{table}

\subsection{Recognition and Speedup Performance}
\label{sec:speedup}

Table \ref{tab:wer} summarizes the WERs for the three ADPSGD strategies using 16, 32 and 64 learners. The total batch size is 8,192 segments. All three ADPSGD variants use the same learning rate schedule: a 10-epoch warmup to 3.2 followed by annealing at $\frac{1}{\sqrt{2}}$ per epoch afterwards. The WERs are reported at epoch 16 with the corresponding training time. Each learner has 4 I/O data loading processes on CPU which overlap with the gradient evaluation on GPU to speed up training. It can be seen that training is significantly accelerated compared to the single-GPU setting. Note that D1D is implemented using NCCL, which is highly optimized, while FM and RM are implemented with MPI for partial model averaging. This also contributes to the longer training time of FM and RM versus D1D. The MPI implementation can be further optimized. Batch sizes larger than 8,192 segments will give us further speedup but incur unacceptable WER degradation.

\begin{table*}[hbt]
 \centering
\begin{tabular}{c|c|c|c|c|c|c|c|c|c|c}
\hline
\multirow{2}{*}{} & Single Learner  & \multicolumn{3}{c|}{16 Learners} & \multicolumn{3}{c|}{32 Learners} & \multicolumn{3}{c}{64 Learners} \\ \cline{3-11}
                  & (batch size 256)  & FM     &  RM     & D1D    & FM    & RM     & D1D    & FM    & RM     & D1D     \\ \hline\hline
SWB(\%)           & 7.5       & 7.6    &  7.6    & 7.4    & 7.9   & 7.7    & 7.6    & 8.1   & 7.8    & 7.5    \\ \hline
CH(\%)            & 13.0      & 13.2   &  13.1   & 13.3   & 13.6  & 13.4   & 13.1   & 14.0  & 13.4   & 13.3   \\ \hline\hline
time(h)           & 99.00     & 15.24  &  15.19  & 5.88   & 8.84  & 8.79   & 3.60   & 5.08  & 5.00   & 2.28   \\ \hline
\end{tabular}
\vspace{0.3cm}
\caption{WER comparison after 16 epochs for fixed mixing, random mixing, and D1D using 16, 32 and 64 learners.}\label{tab:wer}
\end{table*}

\subsection{Slow learners}
\label{sec:straggler}

\begin{figure}[htb]
  \centering
  \centerline{\includegraphics[width=8cm, height=6cm]{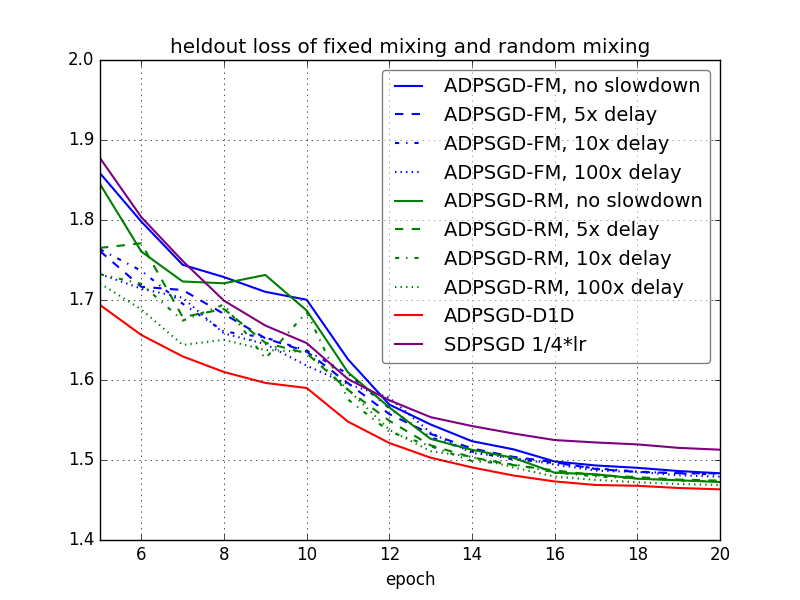}}
  \caption{Convergence of three ADPSGD variants with a straggler that is slowed down by 5x, 10x and 100x, respectively. 8,192-segment batches and 16 learners are used.}\label{fig:straggler}
\end{figure}

Table~\ref{tab:straggler} shows the runtime and recognition performance of the three ADPSGD variants when one learner is a straggler. In this experiment, we purposely slow down one of the learners. We estimate the time taken to finish one batch and make that learner sleep for a pre-defined period of time to control the degree of delay. We investigate scenarios in which one learner is slower than the others by a factor of 5, 10 and 100 times, and compare the training time and WERs to no-straggler baselines. In the table, the ``slowdown'' columns show the ratio between the time needed to finish one epoch with a straggler and the time with no stragglers. As can be observed from the table, ADPSGD-FM and ADPSGD-RM are insensitive to stragglers. Even in the extreme case when one learner slows down by 100 times, there is only a 1.3x and 1.2x slowdown in the training time for ADPSGD-FM and ADPSGD-RM, respectively. Meanwhile, the runtime performance of ADPSGD-D1D, which relies on synchronization, quickly deteriorates. SDPSGD will have similar runtime performance as APSGD-D1D in this case.  When a straggler is present, ADPSGD-FM and ADPSGD-RM can finish training almost as fast as when there are no stragglers and with only a negligible degradation of WERs. Fig.\ref{fig:straggler} compares the convergence curves of ADPSGD-FM (blue) and ADPSGD-RM (green) to those of ADPSGD-D1D and SDPSGD under various straggler conditions and shows that the FM and RM variants converge similarly in straggler and no-straggler conditions.

\begin{table*}[tbh]
\centering
\begin{tabular}{l|c|c|c|c|c|c|c|c|c} \hline
\multirow{3}{*}{slow learner} & \multicolumn{3}{c|}{ADPSGD-FM}  & \multicolumn{3}{c|}{ADPSGD-RM} & \multicolumn{3}{c}{ADPSGD-D1D}  \\ \cline{2-10}
& \multirow{2}{*}{slowdown}   &  \multicolumn{2}{c|}{WER}  &  \multirow{2}{*}{slowdown}   &  \multicolumn{2}{c|}{WER}  &  \multirow{2}{*}{slowdown}   &   \multicolumn{2}{c}{WER}  \\ \cline{3-4}\cline{6-7}\cline{9-10}
             &         &    SWB    &   CH    &          &   SWB    &   CH    &          &    SWB    &     CH   \\ \hline
no delay     &  1.0x   &    7.6    &  13.2   &   1.0x   &   7.6    &  13.1   &   1.0x   &    7.4    &   13.3   \\ \hline
5x delay     &  1.1x   &    7.5    &  13.2   &   1.1x   &   7.5    &  13.5   &   4.1x   &    7.4    &   13.3   \\ \hline
10x delay    &  1.1x   &    7.7    &  13.2   &   1.1x   &   7.6    &  13.0   &   8.7x   &    7.4    &   13.3   \\ \hline
100x delay   &  1.3x   &    7.6    &  13.4   &   1.2x   &   7.6    &  13.2   &  91.7x   &    7.4    &   13.3   \\ \hline
\end{tabular}
\vspace{0.3cm}
\caption{Runtime and WER measurements for three ADPSGD variants using 8,192-segment batches and 16 GPUs when one GPU slows down by 5, 10 and 100 times. The ``slowdown'' columns show the ratio of time taken to finish one epoch with a straggler to the time taken with no stragglers.}\label{tab:straggler}
\end{table*}

\subsection{Scaling Performance of D1D}
\label{sec:d1d}

When all learners are equally fast, the previous experiments show that ADPSGD-D1D gives the best scaling performance. In this experiment, we further test the limits of fast training on SWB2000. Table \ref{tab:wer_time} presents the WERs and training time of the LSTM acoustic models after 16 epochs. As the baseline, we report single-GPU performance with various batch sizes: it takes 122 hours to train the model with 128-segment batches, 99 hours with 256-segment batches, and 82 hours with 512-segment batches. ADPSGD-D1D is performed using 16, 32, 64 and 128 learners with a fixed total batch size of 8,192 segments. If we choose the model trained using a single GPU with 256-segment batches as the reference model, then it takes 99 hours on the IBM supercomputer to train a model with WERs 7.5\% and 13.0\% on the SWB and CH test sets, respectively. Using ADPSGD-D1D on 128 GPUS, it takes only 1.98 hours to train a model that still achieves competitive WERs (7.7\% vs. 7.5\%; 13.3\% vs. 13.0 \%). This amounts to a 50 times speedup. To the best of our knowledge, this is the fastest training time on SWB2000 with this level of WERs ever reported in the literature.

\begin{table*}[htb]
\centering
\begin{tabular}{c|c|c|c|c|c|c}\hline
\multirow{2}{*}{Learner(s)} & \multirow{2}{*}{batch size/learner}  &  \multirow{2}{*}{total batch size} & \multicolumn{2}{c|}{WER(\%)}  & \multirow{2}{*}{Time(h)} &  \multirow{2}{*}{speedup}  \\ \cline{4-5}
                   &       &        &   SWB     &     CH            &                   &                  \\ \hline
        1          &  128  &   128  &   7.5     &    13.0           &       121.96      &   --     \\ \hline
        1          &  256  &   256  &   7.5     &    13.0           &        99.00      &   1.0x      \\ \hline
        1          &  512  &   512  &   7.5     &    13.1           &        82.06      &   1.2x      \\ \hline
        16         &  512  & 8,192  &   7.4     &    13.3           &        5.88       &   16.8x      \\ \hline
        32         &  256  & 8,192  &   7.6     &    13.1           &        3.60       &   27.5x      \\ \hline
        64         &  128  & 8,192  &   7.5     &    13.3           &        2.28       &   43.4x      \\ \hline
  \textbf{128}     & \textbf{64}  & \textbf{8,192}  &  \textbf{7.7}     &    \textbf{13.3}           &        \textbf{1.98}   &  \textbf{50.0x}   \\ \hline
\end{tabular}
\vspace{0.3cm}
\caption{WERs and training time for ADPSGD-D1D under batch size 8,192 segments using various numbers of learners. The speedup is measured against batch size 256 segments for consistency. Given the fixed global batch size 8192 segments, local batch size varies under various numbers of learners.  Therefore one may observe speedup is larger than the increase of learners in some conditions.}
\label{tab:wer_time}
\end{table*}

\section{Discussion}
\label{sec:disc}

Compared to the conventional allreduce-based synchronous PSGD, the three ADPSGD strategies investigated in this paper are more advantageous when dealing with large batches, which is a key factor for successful large-scale data parallel distributed training. Among the three, ADPSGD-D1D enjoys the fastest empirical convergence speed as it achieves averaging consensus among the learners in one step. Nevertheless, it may still suffer from the straggler problem. ADPSGD-FM and ADPSGD-RM are resilient to stragglers, but they take more time to reach consensus among the learners. Random mixing, however, can significantly accelerate the convergence over the fixed mixing by randomizing neighbors to communicate in each iteration. These properties are summarized in Table \ref{tab:traffic-analysis}.

\begin{table}[tbh]
\centering
\begin{tabular}{c|c|c}\hline
               & Convergence  &  Straggler Resilience    \\ \hline\hline
ADPSGD-FM      & slow         &    $\checkmark$          \\ \hline
ADPSGD-RM      & medium       &    $\checkmark$          \\ \hline
ADPSGD-D1D     & fast         &    \xmark                \\ \hline
\end{tabular}
\vspace{0.3cm}
\caption{Comparisons of convergence speed and straggler resilience for the three ADPSGD strategies.}\label{tab:traffic-analysis}
\end{table}

\begin{figure}[htb]
  \centering
  \centerline{\includegraphics[width=6.5cm, height=8cm]{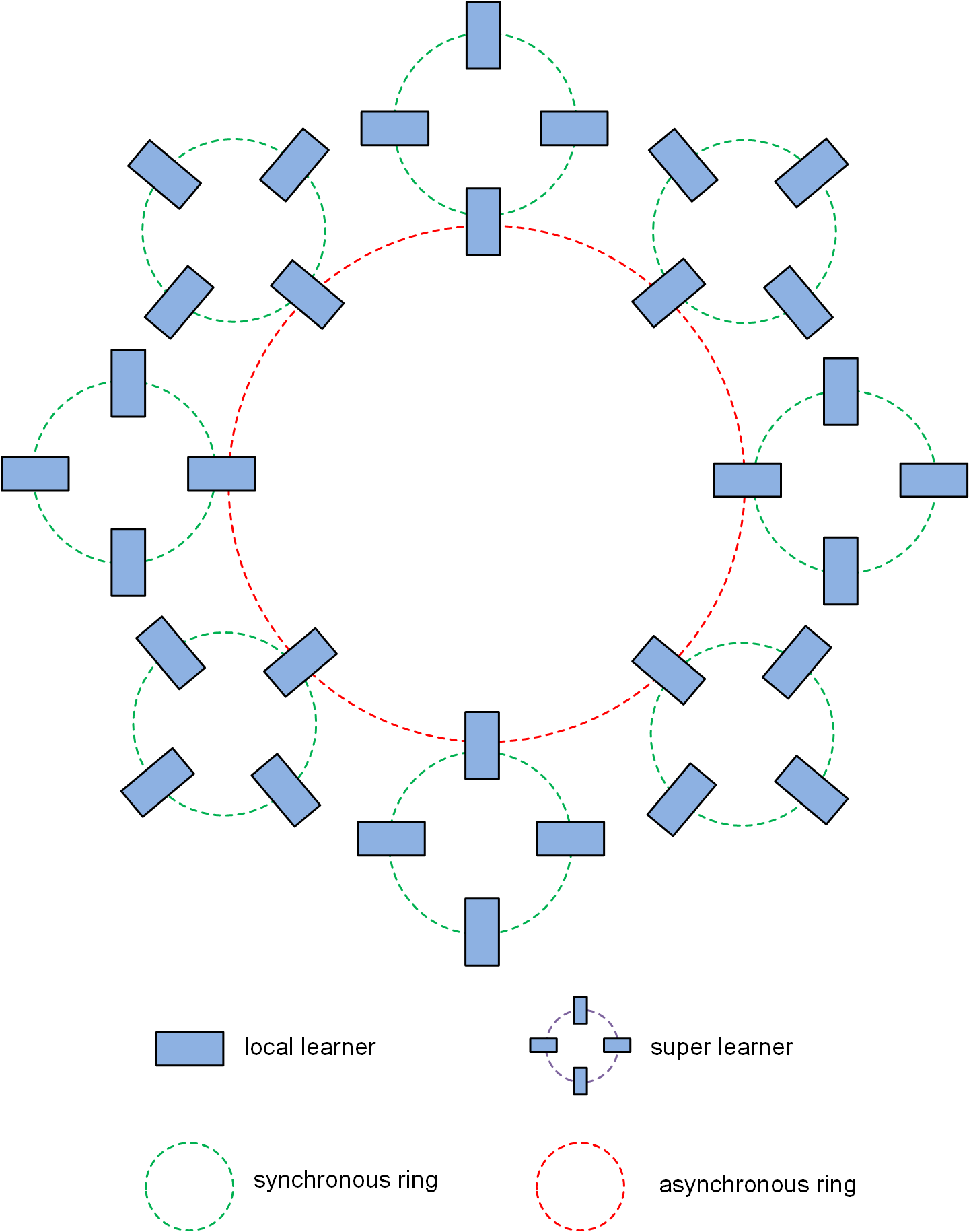}}
  \caption{A hierarchical architecture that combines synchronous and asynchronous training modes where homogeneous learners form a super-learner via local synchronous rings while heterogeneous nodes form a global asynchronous ring.}\label{fig:hring}
\end{figure}

Given their properties, a suitable algorithm can be selected for distributed training depending on the available computing environment. For instance, in a homogeneous computing environment such as the supercomputer used in the above experiments, where even the slowest communication link bandwidth is 25GB/s and all computing devices are highly homogeneous, D1D would be a good choice. On the other hand, when the distributed training has to be carried out in a cloud environment with heterogenous computing devices where the straggler issue becomes more prominent, an algorithm built on a global barrier such as allreduce could be problematic. Under this condition, ADPSGD with fixed or randomized mixing would be a better choice as they do not rely on global synchronization. It is worth noting that the three ADPSGD strategies do not need to be exclusive and they can be combined to strive for better overall efficiency in distributed training. For example, a hybrid strategy can be used to deal with heterogeneous computing environments. Consider a computing environment where multiple nodes form a cluster. Although computing devices on different nodes may have different computing capabilities (e.g V100 vs. K80 GPUs), they are typically homogenous on the same node. The communication within the node is fast while the communication across nodes is slow. Fig.~\ref{fig:hring} illustrates such a hybrid training strategy proposed in \cite{Zhang_DDLASR} which integrates synchronous and asynchronous modes in one hierarchical architecture. Local homogeneous learners form a synchronous ring (e.g ADPSGD-D1D or conventional synchronous PSGD). They are called super-learners. These super-learners then form another global asynchronous ring among the nodes (e.g. ADPSGD-FM or ADPSGD-RM). The synchronous rings take advantage of fast local consensus while the asynchronous ring avoids global synchronization to speed up the communication among super-learners.

The communication ring implemented in this work is based on CPUs. This is intended for a generic communication framework with various hardware configurations such as NVLink or PCIe. However, if more advanced hardware allows, for instance with the availability of NVLink and Infiniband, direct GPU-to-GPU communication within and across nodes is possible and worth investigating. This can further improve the communication efficiency for even better speedup performance.

Although a hybrid LSTM acoustic model trained with the cross-entropy loss is used as a vehicle to investigate the ADPSGD strategies, the conclusions and comparisons can also be extended in principle to sequence training or end-to-end acoustic models, which are not the focus of this paper.

\section{Summary}
\label{sec:sum}

In this paper we introduce three ADPSGD strategies for large-scale distributed acoustic model training for ASR: ADPSGD with fixed mixing, random mixing and delay-by-one communication. We study their theoretical and empirical convergence behaviors and discuss their pros and cons in various computing environments. It is shown that ADPSGD algorithms are more tolerant of large batch sizes than synchronous PSGD, which is important for scaling out.  Recognition and speedup performance are evaluated on the SWB2000 dataset for training of LSTM-based acoustic models under a variety of training configurations. We show that ADPSGD-FM and ADPSGD-RM can give robust runtime and recognition performance even when a straggler is present. When learners are equally fast, ADPSGD-D1D performs very well with large batch sizes. In particular, we show that an LSTM acoustic model can be trained on 2000 hours of data in less than 2 hours on an IBM POWER9 supercomputer with competitive WERs.

\bibliographystyle{IEEETran}
\bibliography{adpsgd}

\appendices
\section{The Spectral Gap of ADPSGD with Fixed Mixing}
\label{app:specgap_fm}

Notice that the mixing matrix $\mathbf{T}^{f}_{k}$ in Eq.~\ref{eqn:matrix_fm} is not only doubly stochastic but also circulant. Its eigenvalues are simply the DFT of its first row \cite{Gray_Toeplitz}:
\begin{align}
     \lambda_{k} & =  \sum_{l=0}^{L-1}t_{1l}e^{-i\frac{2\pi lk}{L}} \nonumber \\
                 & =  \frac{1}{3} + \frac{1}{3}e^{-i\frac{2\pi k}{L}} + \frac{1}{3}e^{-i\frac{2\pi (L-1)}{L}} \nonumber \\
                 & =  \frac{1}{3} + \frac{2}{3}\cos\left(\frac{2\pi k}{L}\right)
\end{align}
It is trivial to see that the largest eigenvalues is 1 which is obtained when $k=0$ while the second largest eigenvalue in magnitude is
\begin{align}
   \hat{\lambda} = \frac{1}{3} + \frac{2}{3}\cos\left(\frac{2\pi}{L}\right).
\end{align}

\section{The Spectral Gap of ADPSGD with Random Mixing}
\label{app:specgap_rm}

Given the random mixing matrix at iteration $k$
\begin{align}
    \mathbf{T}^{r}_{k} = \trans{\mathbf{P}}_{k}\mathbf{T}^{f}_{k}\mathbf{P}_{k},   \label{eqn:randperm_appendix}
\end{align}
taking expectation over the random permutation matrix $\mathbf{P}_{k}$, we have
\begin{align}
\widetilde{\mathbf{T}^{r}_{k}} = \mathbf{E}_{\sigma}[\trans{\mathbf{T}^{r}_{k}}\mathbf{T}^{r}_{k}] = \mathbf{E}_{\sigma}[\trans{\mathbf{P}}_{k}\trans{\mathbf{T}^{f}_{k}}\mathbf{T}^{f}_{k}\mathbf{P}_{k}].
\end{align}
Write $\mathbf{P}_{k}=[p_{1},p_{2},\cdots,p_{L}]$ where $p_{i}$\footnote{The subscript $k$ is dropped here to avoid cluttered notation.} are the columns of $\mathbf{P}_{k}$. Then
\begin{align}
   \widetilde{\mathbf{T}^{r}_{k}}_{,ij} & = \mathbf{E}_{\sigma}[\trans{p}_{i}\trans{\mathbf{T}^{f}_{k}}\mathbf{T}^{f}_{k}p_{j}] \nonumber \\
     & = \text{tr}(\trans{\mathbf{T}^{f}_{k}}\mathbf{T}^{f}_{k}\mathbf{E}_{\sigma}[p_{j}\trans{p}_{i}]) \nonumber \\
     & = \text{tr}(\trans{\mathbf{T}^{f}_{k}}\mathbf{T}^{f}_{k}\mathbf{C}_{ji})  \label{eqn:expperm}
\end{align}
where $\mathbf{C}_{ij}$ is the cross-correlation $p_{i}$ and $p_{j}$.

Given the definition of $\mathbf{T}^{f}_{k}$ in Eq.~\ref{eqn:matrix_fm}, we have\footnote{We assume $L>5$ is reasonably large  }
\begin{align}
    \trans{\mathbf{T}^{f}_{k}}\mathbf{T}^{f}_{k} =   \begin{bmatrix}
    \frac{3}{9} & \frac{2}{9} & \frac{1}{9} &  0   &  \cdots &  0  &  \frac{1}{9} & \frac{2}{9} \\
    \frac{2}{9} & \frac{3}{9} & \frac{2}{9} & \frac{1}{9}  & 0 &  \cdots  &  0 & \frac{2}{9} \\
    \vdots & \vdots & \vdots & \vdots  & \vdots &  \vdots  & \vdots & \vdots  \\
    \frac{2}{9} & \frac{1}{9} & 0 &  0   &  \cdots &  \frac{2}{9}  &  \frac{2}{9} & \frac{3}{9}
\end{bmatrix}\label{eqn:tt}
\end{align}
Next, by taking a look at $\mathbf{E}_{\sigma}[p_{j}\trans{p}_{i}]$, we have
\begin{align}
    \mathbf{E}_{\sigma}[p_{j}\trans{p}_{i}] =  \text{diag}\left\{\frac{1}{L},\cdots,\frac{1}{L}\right\}, \ \ \ \ \text{if} \ i = j  \label{eqn:pp1}
\end{align}
and
\begin{align}
    \mathbf{E}_{\sigma}[p_{j}\trans{p}_{i}] = &  \frac{\mathbf{1}_{\scriptit{L}}\trans{\mathbf{1}}_{\scriptit{L}}}{L(L-1)} - \nonumber \\
     & \text{diag}\left\{\frac{1}{L(L-1)},\cdots,\frac{1}{L(L-1)}\right\}, \ \ \ \ \text{if} \ i\neq j.  \label{eqn:pp2}
\end{align}
Substituting Eq.~\ref{eqn:tt}, Eq.~\ref{eqn:pp1} and Eq.~\ref{eqn:pp2} back to Eq.~\ref{eqn:expperm}, we have
\begin{align}
    \widetilde{\mathbf{T}^{r}_{k}}_{,ii} = \frac{1}{3}, \ \ \widetilde{\mathbf{T}^{r}_{k}}_{,ij} = \frac{2}{3(L-1)}, \ i \neq j   \label{eqn:expperm_elem}
\end{align}

With Eq.~\ref{eqn:expperm} and Eq.~\ref{eqn:expperm_elem} and let $\mathbf{T}_{u}=\frac{\mathbf{1}_{L}\trans{\mathbf{1}}_{L}}{L}$ we have
\begin{align}
  	  & \mathbf{E}\left\|\mathbf{T}^{r}_{1}\mathbf{T}^{r}_{2}\cdots\mathbf{T}^{r}_{k}-\mathbf{T}_{u}\right\|^{2}_2  \nonumber\\
\leq  & \mathbf{E}\left\|\mathbf{T}^{r}_{1}\mathbf{T}^{r}_{2}\cdots\mathbf{T}^{r}_{k}-\mathbf{T}_{u}\right\|^{2}_{F} \nonumber\\
=     &  \mathbf{E}\left[\text{tr}\left(\trans{\mathbf{T}^{r}_k}\ldots\trans{\mathbf{T}^{r}_1}\mathbf{T}^{r}_1\ldots\mathbf{T}^{r}_k\right)\right]      - \mathbf{E}\left[\text{tr}\left(\trans{\mathbf{T}^{r}_k}\ldots\trans{\mathbf{T}^{r}_1}\mathbf{T}_{u}\right)\right] \nonumber \\
      & - \mathbf{E}\left[\text{tr}\left(\trans{\mathbf{T}}_{u}\mathbf{T}^{r}_1\ldots\mathbf{T}^{r}_{k}\right)\right]  + \mathbf{E}\left[\text{tr}\left(\trans{\mathbf{T}}_{u}\mathbf{T}_{u}\right)\right]  \nonumber \\
=  & \text{tr}\left(\mathbf{E}[\trans{\mathbf{T}^{r}_k}\ldots\trans{\mathbf{T}^{r}_1}\mathbf{T}^{r}_1\ldots\mathbf{T}^{r}_k]\right)-1-1+1 \nonumber \\
=  & \text{tr}\left(\left\{\widetilde{\mathbf{T}^{r}_{1}}\right\}^{k}\right)-1
\end{align}
From Eq.~\ref{eqn:expperm_elem}, it can be shown that the eigenvalues of $\widetilde{\mathbf{T}^{r}_{1}}$ are
\begin{align}
  \left\{1, \  \frac{1}{3}-\frac{2}{3(L-1)}, \ \cdots, \  \frac{1}{3}-\frac{2}{3(L-1)}\right\}
\end{align}
accordingly, the eigenvalues of $\left\{\widetilde{\mathbf{T}^{r}_{1}}\right\}^{k}$ are
\begin{align}
  \left\{1, \  \left(\frac{1}{3}-\frac{2}{3(L-1)}\right)^{k}, \ \cdots, \  \left(\frac{1}{3}-\frac{2}{3(L-1)}\right)^{k}\right\}
\end{align}
It follows that
\begin{align}
  	  & \mathbf{E}\left\|\mathbf{T}^{r}_{1}\mathbf{T}^{r}_{2}\cdots\mathbf{T}^{r}_{k}-\mathbf{T}_{u}\right\|^{2}_2  \nonumber\\
\leq  &  \text{tr}\left(\left\{\widetilde{\mathbf{T}^{r}_{1}}\right\}^{k}\right)-1  \nonumber \\
  =   & 1 + (L-1)\left(\frac{1}{3}-\frac{2}{3(L-1)}\right)^{k}-1 \nonumber \\
  =   & (L-1)\left(\frac{1}{3}-\frac{2}{3(L-1)}\right)^{k} \nonumber   \\
\leq  & \frac{L-1}{3^{k}}
\end{align}
which gives
\begin{align}
\mathbf{E}\left\|\mathbf{T}^{r}_{1}\mathbf{T}^{r}_{2}\cdots\mathbf{T}^{r}_{k}-\frac{\mathbf{1}_{L}\trans{\mathbf{1}}_{L}}{L}\right\|_2 \leq \frac{\sqrt{L-1}}{(\sqrt{3})^{k}}.
\end{align}

\end{document}